\RequirePackage{fix-cm}

\documentclass[twocolumn]{svjour3}          
\smartqed  

\usepackage{listings}
\usepackage{color}

\definecolor{dkgreen}{rgb}{0,0.6,0}
\definecolor{gray}{rgb}{0.5,0.5,0.5}
\definecolor{mauve}{rgb}{0.58,0,0.82}

\usepackage{natbib}
\usepackage{graphicx}
\usepackage{amsmath}
\usepackage{xcolor}
\usepackage[breaklinks=true]{hyperref}
\usepackage{breakcites}
\usepackage{url}
\journalname{Soft Computing}
\def\makeheadbox{\relax}
\begin{document}

\title{On the Effects of Pseudo and Quantum Random Number Generators in Soft Computing
}
\subtitle{}


\author{Jordan J. Bird        \and
        Anik\'o Ek\'art \and
        Diego R. Faria 
}


\institute{Jordan J. Bird \at
              School of Engineering and Applied Science \\
              Aston University\\
              \email{birdj1@aston.ac.uk}           
\and
           Anik\'o Ek\'art \at
              School of Engineering and Applied Science \\
              Aston University\\
              \email{d.faria@aston.ac.uk}
           \and
           Diego R. Faria \at
              School of Engineering and Applied Science \\
              Aston University\\
              \email{d.faria@aston.ac.uk}
          }

\date{Received: date / Accepted: date}

\maketitle

\begin{abstract}
In this work, we argue that the implications of Pseudo and Quantum Random Number Generators (PRNG and QRNG) inexplicably affect the performances and behaviours of various machine learning models that require a random input. These implications are yet to be explored in Soft Computing until this work. We use a CPU and a QPU to generate random numbers for multiple Machine Learning techniques. Random numbers are employed in the random initial weight distributions of Dense and Convolutional Neural Networks, in which results show a profound difference in learning patterns for the two. In 50 Dense Neural Networks (25 PRNG/25 QRNG), QRNG increases over PRNG for accent classification at +0.1\%, and QRNG exceeded PRNG for mental state EEG classification by +2.82\%. In 50 Convolutional Neural Networks (25 PRNG/25 QRNG), the MNIST and CIFAR-10 problems are benchmarked, in MNIST the QRNG experiences a higher starting accuracy than the PRNG but ultimately only exceeds it by 0.02\%. In CIFAR-10, the QRNG outperforms PRNG by +0.92\%. The n-random split of a Random Tree is enhanced towards and new Quantum Random Tree (QRT) model, which has differing classification abilities to its classical counterpart, 200 trees are trained and compared (100 PRNG/100 QRNG). Using the accent and EEG classification datasets, a QRT seemed inferior to a RT as it performed on average worse by -0.12\%. This pattern is also seen in the EEG classification problem, where a QRT performs worse than a RT by -0.28\%. Finally, the QRT is ensembled into a Quantum Random Forest (QRF), which also has a noticeable effect when compared to the standard Random Forest (RF). 10 to 100 ensembles of Trees are benchmarked for the accent and EEG classification problems. In accent classification, the best RF (100 RT) outperforms the best QRF (100 QRF) by 0.14\% accuracy. In EEG classification, the best RF (100 RT) outperforms the best QRF (100 QRT) by 0.08\% but is extremely more complex, requiring twice the amount of trees in committee. All differences are observed to be situationally positive or negative and thus are likely data dependent in their observed functional behaviour. 

\keywords{Quantum Computing \and Soft Computing \and Machine Learning \and Neural Networks \and Classification}

\end{abstract}

\section{Introduction}
\label{intro}
Quantum and Classical hypotheses of our reality are individually definitive and yet are independently paradoxical, in that they are both scientifically verified though contradictory to one another. These concurrently antithetical, nevertheless infallible natures of the two models have enflamed debate between researchers since the days of Albert Einstein and Erwin Schr{\"o}dinger during the early 20$^{th}$ century. Though the lack of a \textit{Standard Model of the Universe} continues to provide a problem for physicists, the field of Computer Science thrives by making use of both in Classical and Quantum computing paradigms since they are independently observable in nature. \\

Though the vast majority of computers available are classical, Quantum Computing has been emerging since the late 20$^{th}$ Century, and is becoming more and more available for use by researchers and private institutions. Cloud platforms developed by industry leaders such as Google, IBM, Microsoft and Rigetti are quickly growing in resources and operational size. This rapidly expanding availability of quantum computational resources allows for researchers to perform computational experiments, such as heuristic searches or machine learning, but allow for the use of the laws of quantum mechanics in their processes. For example, for \textit{n} computational bits in a state of entanglement, only one needs to be measured for all \textit{n} bits to be measured, since they all exist in parallel or anti-parallel relationships. Through this process, computational complexity has been reduced by a factor of \textit{n}. Bounded-error Quantum Polynomial time (BQP) problems are a set of computational problems which cannot be solved by a classical computer in polynomial time, whereas a quantum processor has the ability to with its different laws of physics.  \\

Optimisation is a large multi-field conglomeration of research, which is rapidly accelerating due to the growing availability of powerful computing hardware such has CUDA. Examples include Ant Colony Optimisation inspired by the pheromone-dictated behaviour of ants \cite{deng2019improved}, orthoganal translations to derive a Principle Component Analysis \cite{zhao2019fault}, velocity-based searches of particle swarms \cite{deng2017study}, as well as entropy-based methods of data analysis and classification \cite{zhao2018study}.

There are several main contributions presented by this research:
\begin{enumerate}
\item A comparison of the abilities of Dense Neural Networks with their initial random weight distributions derived by Pseudorandom and Quantum Random methods.
\item An exploration of Random Tree models compared to \textit{Quantum Random Tree} models, which utilise Pseudorandom and Quantum Random Number Generators in their generation respectively.
\item A benchmark of the number of Random Trees in a Random Forest model compared to the number of Quantum Random Trees in a Quantum Random Forest model.
\item A comparison of the effects of Pseudo and True randomness in initial random weight distributions in Computer Vision, applied to Deep Neural Networks and Convolutional Neural Networks.
\end{enumerate}
Although Quantum, Quantum-inspired, and Hybrid Classical/Quantum algorithms are explored, as well as the likewise methods for computing, the use of a Quantum Random Number Generator is rarely explored within a classical machine learning approach in which an RNG is required  ~\cite{kretzschmar2000quantum}. 

This research aims to compare approaches for random number generation in Soft Computing for two laws of physics which directly defy one another; the Classical \textit{true randomness is impossible} and the Quantum \textit{true randomness is possible}  ~\cite{calude2008quantum}. Through the application of both Classical and Quantum Computing, simulated and true random number generation are tested and compared via the use of a Central Processing Unit (CPU) and an electron spin-based Quantum Processing Unit (QPU) via placing the subatomic particle into a state of quantum superposition. Logic would conjecture that the results between the two ought to be indistinguishable from one another, but experimentation within this study suggests otherwise. The rest of this article is structured as follows:

Section \ref{Background} gives an overview of the background to this project and important related theories and works. Specifically, Quantum Computing, the differing ideas of randomness in both Classical and Quantum computing, applications of quantum theory in computing and finally a short subsection on the machine learning theories used in this study. Section \ref{experimentalsetup} describes the configuration of the models as well as the methods used specifically to realise the scientific studies in this article, before being presented and analysed in Section \ref{resultsdiscussion}. The experimental results are divided into four individual experiments:
\begin{itemize}
\item Experiment 1 - On random weight distribution in Dense Neural Networks: Pseudorandom and Quantum Random Number Generators are used to initialise the weights in Neural Network models. 
\item Experiment 2 - On Random Tree splits: The \textit{n} Random Splits for a Random Tree classifier are formed by Pseudo and Quantum Random numbers. 
\item Experiment 3 - On Random Tree splits in Random Forests: The \textit{Quantum Tree} model derived from Experiment 2 is used in a \textit{Quantum Random Forest} ensemble classifier.
\item Experiment 4 - On Computer Vision: A Deep Neural Network and Convolutional Neural Network are trained on two image recognition datasets with pseudo and true random weight distributions for the application of Computer Vision.
\end{itemize}
Experiments are separated in order to focus upon the effects of differing random number generators on a specific model. Explored in these are the effects of Pseudorandom and Quantum Random number generation in their processes, and a discussion of similarities and differences between the two in terms of statistics as well as their wider effect on the classification process. Section \ref{futurework} outlines possible extensions to this study for future works, and finally, a conclusion is presented in Section \ref{conclusion}. 

\section{Background and Related Works} \label{Background}
\subsection{Quantum Computing}
Pioneered by Paul Benioff's 1980 work  ~\cite{benioff1980computer}, Quantum Computing is a system of computation that makes computational use of phenomena outside of classical physics such as the entanglement and superposition of subatomic particles  ~\cite{gershenfeld1998quantum}. Whereas classical computing is concerned with electronic bits that have values of 0 or 1 and logic gates to process them, quantum computing uses both classical bits and gates as well as new possible states; such as a bit being in a state of superposition (0 and 1) or entangled with other bits. Entanglement means that the value of the bit, even before measurement, can be assumed to be parallel or anti-parallel to another bit of which it is entangled to  ~\cite{bell1964einstein}. These extended laws allow for the solving of problems far more efficiently than computers. For example, a 64-bit system ($2^{63}-1$) has approximately 9.22 quintillion values with its individual bits at values 1 or 0, whereas unlike a three-state ternary system which QPUs are often mistaken for, the laws superposition and the degrees of state would allow a small array of qubits to represent all of these values at once - theoretically allowing quantum computers to solve problems that classical computers will never be able to possibly solve. Since the stability of entanglement decreases with the more computational qubits used, only very small-scale experiments have been performed as of today. Quantum Processing Units (QPUs) made available for use by Rigetti, Google and IBM have up to 16 available qubits for computing via their cloud platforms. 

\subsection{Randomness in Classical and Quantum Computing}
In classical computing, randomness is not random, rather, it is simulated by a \textit{pseudo-random} process. Processor architectures and Operating Systems have individual methods of generating pseudo-random numbers which must conform to cybersecurity standards such as \textit{NIST}  ~\cite{barker2007recommendation}. Major issues arise with the possibility of \textit{backdoors}, notably for example Intel's pseudo random generator which, after hijacking, allowed for complete control of a computer system for malicious intent  ~\cite{degabriele2016backdoors, schneier2007did}. The Intel issue was far from a lone incident, the RANDU system was cracked by the NSA for unprecedented access to the RSA BSAFE cryptographic library, as well as in 2006 when Debian OpenSSL's random number generator was also cracked, leading to Debian being compromised for two years  ~\cite{markowsky2014sad}. Though there are many methods of deriving a pseudo-random number, all classical methods, due to the laws of classical physics providing limitation, are sourced through arbitrary yet deterministic events  ~\cite{gallego2013full}; such as a combination of, time since \textit{n} last key press, hardware temperature, system clock, lunar calendar etc. This arbitration could possibly hamper or improve algorithms that rely on random numbers, since the state of the executing platform could indeed directly influence their behaviour. \\

According to Bell's famous theorem, \textit{"No physical theory of local hidden variables can ever reproduce all of the predictions of quantum mechanics"}  ~\cite{bell1964einstein}. This directly argued against the position put forward by Einstein et. al in which it is claimed that the Quantum Mechanical 'paradox' is simply due to incomplete theory  ~\cite{einstein1935can}. Using Bell's theorem, demonstrably random numbers can be generated through the fact that observing a particle's state while in superposition gives a true 50/50 outcome (qubit value 0, 1)  ~\cite{pironio2010random}. This concretely random output for the value of the single bit can be used to build integers comprised of larger numbers of bits which, since they are all individually random, their product is too. This process is known as a Quantum Random Number Generator (QRNG).\\

Behaviours in Quantum Mechanics such as, but not limited to, branching path superposition  ~\cite{jennewein2000quantum}, time of arrival  ~\cite{wayne2009photon}, particle emission count  ~\cite{ren2011quantum}, attenuated pulse  ~\cite{wei2009quantum}, and vacuum fluctuations  ~\cite{gabriel2010generator} are all entirely random - and have been used to create true QRNGs. In 2000, it was observed that a true random number generator could be formed through the observation of photons  ~\cite{stefanov2000optical}. Firstly, a beam of light is split into two streams of entangled photons, noise is reduced after which the photons of both streams are observed. The two detectors correlate to 0 and 1 values, and a detection will amend a bit to the result. The detection of a photon is non-deterministic between the two, and therefore a completely random series of values are the result of this experiment.\\ 

This study makes use of the branching path superposition method for the base QRNG, in that the observed state of a particle \textit{c} at time \textit{t}, the state of \textit{c} is non-deterministic until only after observation \textit{t}. In the classical model, the law of superposition simply states that for properties \textit{A} and \textit{B} with outcomes \textit{X} and \textit{Y}, both properties can lead to state \textit{XY}. For example, the translation and rotation of a wheel can lead to a rolling state  ~\cite{cullerne2000penguin}, a third superstate of the two possible states. This translates into quantum physics, where quantum states can be superposed into an additional valid state  ~\cite{dirac1981principles}. \\

   \begin{figure}[]
      \centering
      \includegraphics[scale=0.4]{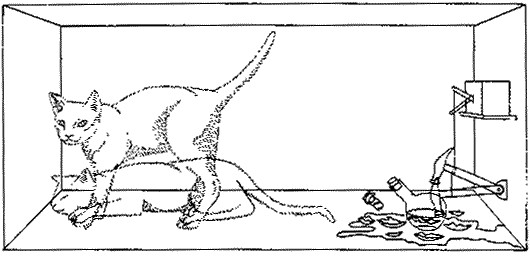}
      \caption{The Famous Schr{\"o}dinger's Cat Thought Experiment. When unobserved, the cat arguably exists in two opposite states (alive and dead), which itself constitutes a third superstate~\cite{schrodinger1935gegenwartige}.}
      \label{cat}
   \end{figure}
   
This is best exemplified with Erwin Schr{\"o}dinger's famous thought experiment, known as \textit{Schr{\"o}dinger's Cat} ~\cite{schrodinger1935gegenwartige}. As seen in Fig. \ref{cat}, a cat sits in a box along with a Geiger Counter and a source of radiation. If alpha radiation is detected, which is a completely random event, the counter releases a poison into the box, killing the cat. The thought experiment explains superposition in such a way, that although the cat has two states (Alive or Dead), when unobserved, the cat is both simultaneously alive and dead. In terms of computing, this means that the two classical behaviours of a single bit, 1 or 0, can be superposed into an additional state, \textit{1 and 0}. \textit{Just as the cat only becomes alive or dead when observed, a superposed qubit only becomes 1 or 0 when measured.}\\

    \begin{figure}[]
      \centering
      \includegraphics[scale=0.4]{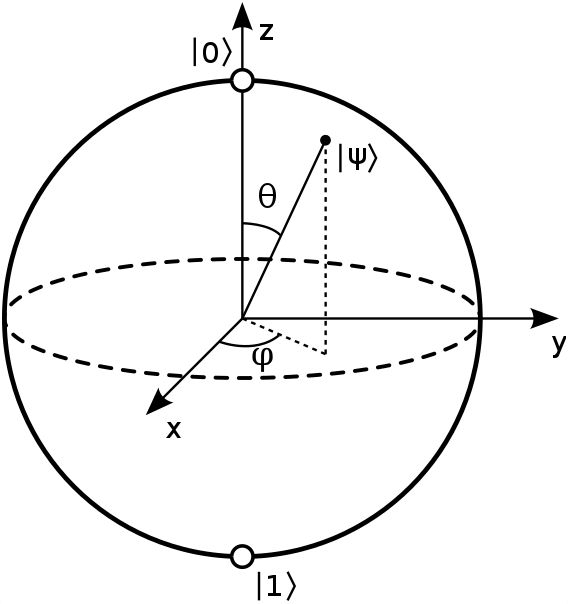}
      \caption{A Bloch Sphere Represents the Two Basis States of a Qubit (0, 1) as well as the States of Superposition In-between.}
      \label{blochsphere}
   \end{figure}

A Bloch Sphere is a graphical representation of a qubit in superposition ~\cite{bloch1946nuclear} and can be seen in Fig. \ref{blochsphere}. In this diagram, the basis states are interpreted by each pole, denoted as $|0\rangle$ and $|1\rangle$. Other behaviours, the rotations of spin about points $\psi$, $\phi$, and $\theta$ are used to superpose the two states to a degree. Thus depending on the method of interpretation, many values can be encoded within only a single bit of memory. \\

The Hadamard Gate within a QPU is a logical gate which coerces a qubit into a state of superposition based on a basis (input) state. 0 is mapped as follows:
\begin{equation}
    |0\rangle \mapsto  \frac{|0\rangle + |1\rangle}{ \sqrt{2} } 
\end{equation}
The other possible basis state, 1, is mapped as:
\begin{equation}
    |0\rangle \mapsto  \frac{|0\rangle - |1\rangle}{ \sqrt{2} } 
\end{equation}
This single qubit quantum Fourier transform is thus represented through the following matrix:
\begin{equation}
    H =  \frac{1}{ \sqrt{2} }   \begin{bmatrix}1 & 1 \\1 & -1 \end{bmatrix} 
\end{equation}

Just as in the thought experiment described in which Schr{\"o}dinger's cat is both alive and dead, the qubit now exists in a state of quantum superposition; it is both 1 and 0. That is, until it is measured, in which there will be an equal probability that the observed state is 1 or 0, giving a completely randomly generated bit value. This is the logical basis of all QRNGs.\\

\subsection{Quantum Theory in Related State-of-the-art Computing Application}
The field of Quantum Computing is young, and thus there are many frontiers of research of which none have been mastered. Quantum theory, though, has been shown in some cases to improve current ideas in Computer Science as well as endow a system with abilities that would be impossible on a classical computer. This section outlines some of the state of the art applications of quantum theory in computing. \\

Quantum Perceptrons are a theoretical approach to deriving a quantum equivalent of a perceptron unit (neuron) within an Artificial Neural Network ~\cite{schuld2014quest}. Current lines of research focus around the possibilities of associative memory through quantum entanglement of internal states within the neurons of the network. The approach is heavily inspired by the notion that the biological brain may operate within both classical and quantum physical space ~\cite{hagan2002quantum}. Preliminary works have found Quantum Neural Networks have a slight statistical advantage over classical techniques within larger and more complex domains ~\cite{narayanan2000quantum}. A very limited extent of research suggest quantum effects in a network to be the possible source of consciousness  ~\cite{hameroff1996orchestrated}, providing an exciting avenue for Artificial Intelligence research in the field of Artificial Consciousness. Inspiration from quantum mechanics has led to the implementation of a Neural Networks based on fuzzy logic systems ~\cite{purushothaman1997quantum}, research showed that QNNs are capable of structure recognition, which sigmoid-activated hidden units within a network cannot. \\

There are many statistical processes that are either more efficient or even simply possible through the use of Quantum Processors. Simon's Problem provides initial proof that there are problems that can be solved exponentially faster when executed in quantum space ~\cite{arora2009computational}. Based on Simon's Problem, Shor's Algorithm uses quantum computing to derive the prime factors of an integer in polynomial time ~\cite{shor1999polynomial}, something which a classical computer is not able to do. \\

Some of the most prominent lines of research in quantum algorithms for Soft Computing are the exploration of Computational Intelligence techniques in quantum space such as meta-heuristic optimisation, heuristic search, and probabilistic optimisation etc. 
Pheromone trails in Ant Colony Optimisation searches generated and measured in the form of qubits with operations of entanglement and superposition for measurement and state scored highly on the \textit{Tennessee Eastman Process} benchmark problem, due to the optimal operations involved ~\cite{wang2007novel}. This work was applied by researchers, who in turn found that combining Support Vector Machines with Quantum Ant Colony Optimisation search provided a highly optimised strategy for solving fault diagnosis problems ~\cite{wang2008novel}, greatly improving the base SVM. Parallel Ant Colony Optimisation has also been observed to greatly improve in performance when operating similar techniques ~\cite{you2010quantum}. Similar techniques have also been used in the genetic search of problem spaces, with quantum logic gates performing genetic operations and probabilistic representations of solution sets in superposition/entanglement, the technique is observed to be superior over its classical counterpart when benchmarked on the combinatorial optimisation problem ~\cite{han2001parallel}.\\

Statistical and Deep Learning techniques are often useful in other scientific fields such as engineering ~\cite{naderpour2019innovative,naderpour2019shear}, medicine ~\cite{khan2001classification,penny1996neural}, chemistry ~\cite{schutt2019unifying,gastegger2019modeling}, and astrophysics ~\cite{krastev2019real,kimmy2019deepcmb} among a great many others ~\cite{carlini2017towards}. As of yet, the applications of quantum solutions have not been applied within these fields towards the possible improvement of soft computing technique.  \\

\section{Experimental Setup and Design} \label{experimentalsetup}
For the generation of true random bit values, an electron-based superposition state is observed using a QPU. The Quantum Assembly Language code for this is given in Appendix A; an electron is transformed using a Hadamard Gate and thus now exists in a state of superposition. When the bit is observed, it takes on a state of either 0 or 1, which is a non-deterministic 50/50 outcome ie. perfect randomness. A VM example of how these operations are formed into a random integer are given in Appendix B; the superposition state particle is sequentially observed and each derived bit is amended to a result until 32 bits have been generated. These 32 bits are then treated as a single binary number. The result of this process is a truly random unsigned 32-bit integer. \\

For the generation of bounded random numbers, the result is normalised with the upper bound being the highest possible value of the intended number. For those that also have lower bounds below zero, a simple subtraction is performed on a higher bound of normalisation to give a range. For example, if a random weight distribution for neural network initialisation is to be generated between -0.5 and 0.5, the random 32-bit integer is normalised between 0-1 and 0.5 is subtracted from the result, giving the desired range. This process is used for the generation of both PRN and QRN since they are therefore then directly comparable with one another and thus also directly relative in their effects upon a machine learning process. \\

For the first dataset in each experiment, a publicly available Accent Classification dataset is retrieved\footnote{https://www.kaggle.com/birdy654/speech-recognition-dataset-england-and-mexico}. This dataset was gathered from subjects from the United Kingdom and Mexico, all speaking the same seven phonetic sounds ten times each. A flat dataset is produced via 27 logs of their Mel-frequency Cepstral Coefficients every 200ms to produce a mathematical description of the audio data. A four-class problem arises in the prediction of the locale of the speaker (West Midlands, London, Mexico City, Chihuahua). The second dataset in each experiment is an EEG brainwave dataset sourced from a previous study ~\cite{jordan2018eegprepublication}\footnote{https://www.kaggle.com/birdy654/eeg-brainwave-dataset-mental-state}. The wave data has been extracted from the TP9, AF7, AF8 and TP10 electrodes, and has been processed in a similar way to the speech in the first dataset, except is done so through a much larger set of mathematical descriptors. For the four-subject EEG dataset, a three-class problem arises; the concentrative state of the subject (concentrating, neutral, relaxed). The feature generation process from this dataset was observed to be effective for mental state classification in the aforementioned study, as well as for emotional classification from the same EEG electrodes ~\cite{jordan2019eegprepublication}.  \\

For the final experiment, two image classification datasets are used. Firstly, the MNIST image dataset is retrieved\footnote{http://yann.lecun.com/exdb/mnist/} ~\cite{lecun-mnisthandwrittendigit-2010} for the MLP. This dataset is comprised of 60,000 32x32 handwritten single digits 0-9, a 10-class problem with each class being that of the digit written. Secondly, the CIFA-10 dataset is retrieved\footnote{https://www.cs.toronto.edu/kriz/cifar.html} ~\cite{cifar10} for a CNN. This, as with the MNIST dataset, is comprised of 60,000 32x32 10-class images of entities (eg. bird, cat, deer). \\

For the generation of pseudorandom numbers, an AMD FX8320 processor is used with given bounds for experiment 1a and 1b. The Java Virtual Machine generates pseudorandom numbers for experiments 2 and 3. All of the pseudorandom number generators had their seed set to the order of execution, ie. the first model has a seed of 1 and the $n^{th}$ model has a seed of $n$. Due to the high resource usage of training a large volume of neural networks, the CUDA cores of an Nvidia GTX980Ti were utilised and they were trained on a 70/30 train/test split of the datasets. For the Machine Learning Models explored in Experiments 2 and 3, 10-fold cross validation was used due to the availability of computational resources to do so. \\

\subsection{Experimental Process}
In this subsection, a step-by-step process is given describing how each model is trained towards comparison between PRNG and QRNG methods. MLP and CNN RNG methods are operated through the same technique and as such are described together, following this, the Random Tree (RT) and Quantum Random Tree (QRT) are described. Finally the ensembles of the two types of trees are then finally described as Random Forest (RF) and Quantum Random Forest (QRF). Each set of models is tested and compared for two different datasets, as previously described. For replicability of these experiments, the code for Random Bit Generation is given in Appendix A (for construction of an n-bit integer). Construction of the n-bit integer through electron observation loop is given in Appendix B. \\

For the Random Neural Networks, all use the ADAM Stochastic Optimiser for weight tuning~\cite{kingma2014adam}, and the activation function of all hidden layers is ReLU~\cite{agarap2018deep}. For Random Trees, $K$ randomly chosen attributes is defined below (acquired via either PRNG or QRNG) and the minimum possible value for $k$ is 1, no pruning is performed. Minimum class variance is set to $-inf$ since the datasets are well-balanced, the maximum depth of the tree is not limited, and classification must always be performed even if confusion occurs. The chosen Random Tree attributes are also used for all trees within Forests, where the random number generator for selection of data subsets is also decided by a PRNG or QRNG. The algorithmic complexity for a Random Tree is given as $O( v \times n log(n) )$ where $n$ is the number of data objects in the dataset and $v$ is the number of attributes belonging to a data object in the set. Algorithmic complexity of the neural networks are dependent on chosen topologies for each problem, and the complexity is presented as an $O(n^2)$ problem. 

Given $n$ number of networks to be benchmarked for $x$ epochs, generally, the MLP and CNN experiments are automated as follows:
\begin{enumerate}
    \item Initialise $n/2$ neural networks with initial random weights generated by an AMD CPU (pseudorandom).
    \item Initialise $n/2$ neural networks with initial random weights generated by a Rigetti QPU (true random).
    \item Train all $n$ neural networks.
    \item Consider classification accuracy at each epoch\footnote{Accuracy/epoch graphs are given in Section \ref{resultsdiscussion}} for comparison as well as statistical analysis of all $n/2$ networks.
\end{enumerate}

Given $n$ number of trees with a decision variable $K_{x}$ ($K$ randomly chosen attributes at node $x$), the process of training Random Trees (RT) and Quantum Random Trees (QRT) are given as follows:
\begin{enumerate}
    \item Train $n/2$ Random Trees, in which the RNG for deciding set $K$ for every $x$ is executed by an AMD CPU (pseudorandom)
    \item Train $n/2$ Quantum Random Trees, in which the RNG for deciding set $K$ for every $x$ is executed by a Rigetti QPU (true random).
    \item Considering the best and worst models, as well as the mean result, compare the two sets of $n/2$ models in terms of statistical difference\footnote{Box and whisker comparisons given in Section \ref{resultsdiscussion}.}
\end{enumerate}

Finally, the Random Tree and Quantum Random Tree are benchmarked as an ensemble, through Random Forests and Quantum Random Forests. This is performed mainly due to the unpruned Random Tree likely overfitting to training data~\cite{hastie2005elements}. The process is as follows\footnote{For further detail on the Random Decision Forest classifier selected for this study, please refer to ~\cite{Breiman2001}}:
\begin{enumerate}
    \item For the Random Forests, benchmark 10 forests containing \{10, 20, 30 ... 100\} Random Tree Models (as generated in the \textit{Random Tree Experimental Process} list above). 
    \item For the Quantum Random Forests, benchmark 10 forests containing \{10, 20, 30 ... 100\} Quantum Random Tree Models (as generated in the \textit{Random Tree Experimental Process} list above). 
    \item Compare abilities of all 20 models, in terms of classification ability as well as the statistical differences, \textit{if any}, between different numbers of trees in the forest.
\end{enumerate}

\section{Results and Discussion} \label{resultsdiscussion}
In this section, results are presented and discussed for multiple Machine Learning models when their random number generator is either Pseudo-randomly, or True (Quantum) Randomly generated. Please note that in neural network training, lines do not correlate on a one-to-one basis. Each line is the accuracy of a neural network throughout the training process, and line colour defines how that network had its weights initialised ie. whether or not it has pseudo or quantum random numbers as its initial weights. 

\subsection{MLP: Random Initialisation of Dense Neural Network Weights}
For Experiment 1, a total of fifty dense neural networks were trained for each dataset. All networks were identical except for their initial weight distributions. Initial random weights within bounds of -0.5 and 0.5 were set, 25 of the networks derived theirs from a PRNG, and the other 25 from a QRNG.
\subsubsection{Accent Classification}
\begin{figure}
    \centering
    \includegraphics[scale=0.55]{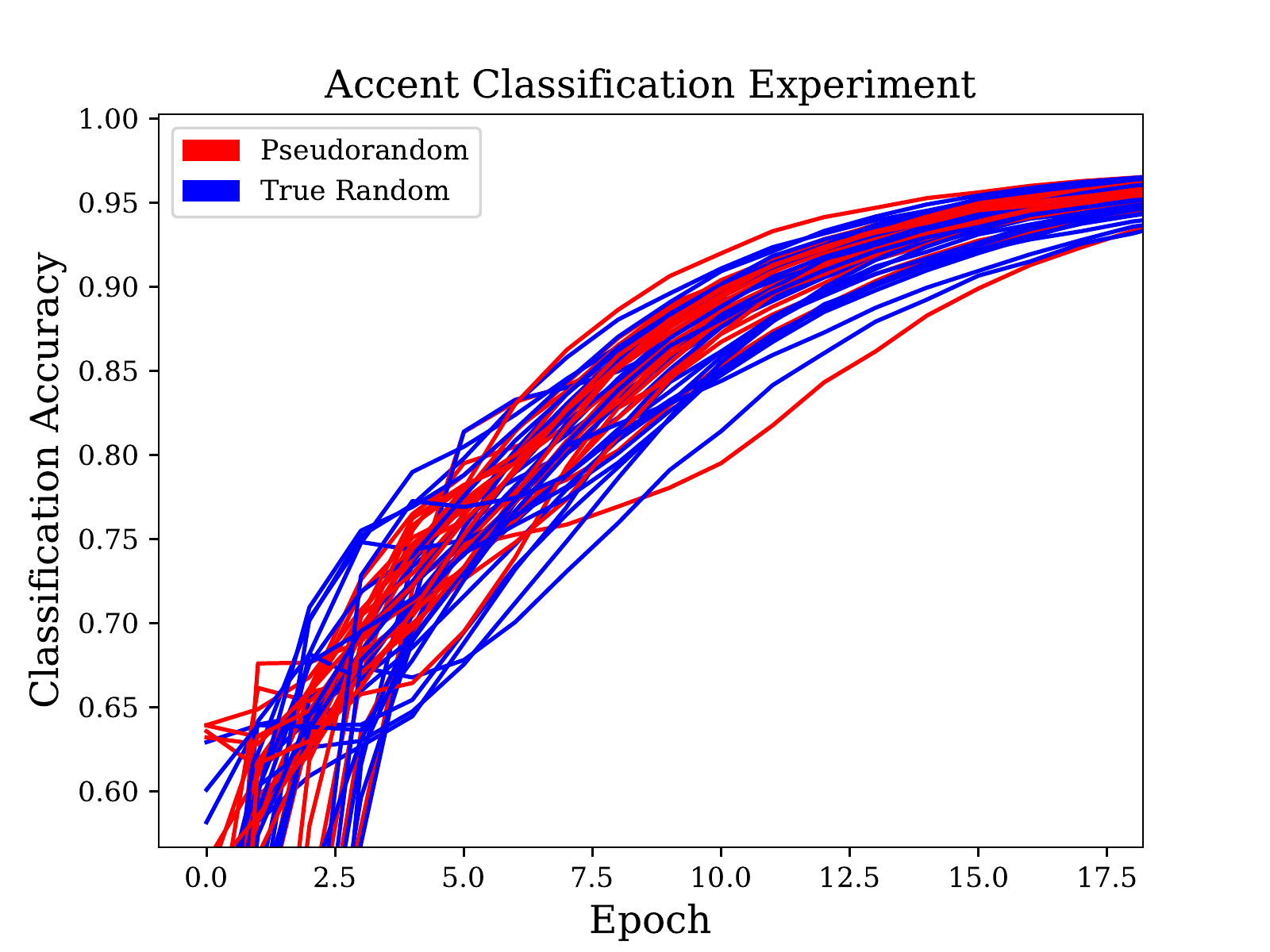}
    \caption{The Main Learning Curve Experienced for 50 Dense Neural Networks, 25 with PRNG and 25 with QRNG Initially Distributed Weights in Accent Classification}
    \label{curvegraphaccent}
\end{figure}

For Experiment 1a, the accent classification dataset was used. In this experiment, we observed initial sparse learning processes before stabilisation occurs at approximately epoch 30 and the two converge upon a similar result. Fig. \ref{curvegraphaccent} shows this convergence of the learning processes the initial learning curve experienced during the first half of the process, in this graph it can be observed that the behaviour of pseudorandom weight distribution is far less erratic than that of the quantum random number generator. This shows that the two methods of random number generators do have an observable effect on the learning processes of a neural network. 

For PRNG, the standard deviation between all 25 final results was 0.00098 suggesting that a classification maxima was being converged upon. The standard deviation for QRNG was considerably larger, but statistically minimal at 0.0017. Mean final results were 98.73\% for PRNG distributions and 98.8\% for QRNG distributions. The maximum classification accuracy achieved by the PRNG initial distribution was 98.8\% whereas QRNG achieved a slightly higher result of 98.9\% at epoch 49. For this problem, the differences between the initial distribution of PRNG and QRNG are minimal, QRNG distribution results are somewhat more entropic than PRNG but otherwise the two sets of results are indistinguishable from one another, and most likely simply due to random noise.

\subsubsection{Mental State Classification}

\begin{figure}
    \centering
    \includegraphics[scale=0.55]{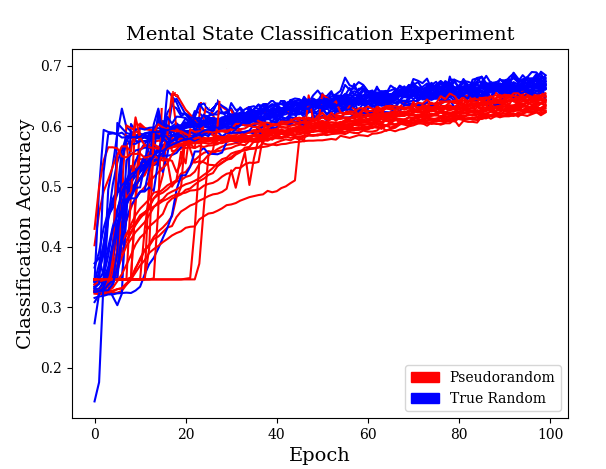}
    \caption{The Full Learning Process of 50 Dense Neural Networks, 25 with PRNG and 25 with QRNG Initially Distributed Weights in Mental State EEG Classification}
    \label{fullgraphmental}
\end{figure}

\begin{figure}
    \centering
    \includegraphics[scale=0.55]{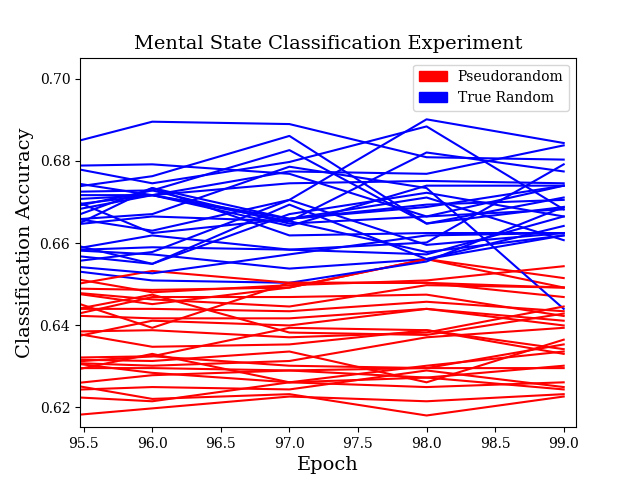}
    \caption{The Final Epochs of Learning for 50 Dense Neural Networks, 25 with PRNG and 25 with QRNG Initially Distributed Weights in Mental State EEG Classification}
    \label{endgraphmental}
\end{figure}

 For Experiment 1b, the Mental State EEG classification dataset was used ~\cite{jordan2018eegprepublication}. Fig. \ref{fullgraphmental} shows the full learning process of the networks from initial epoch 0 up until backpropagation epoch 100, though this graph is erratic and crowded, the emergence of a pattern becomes obvious within epochs 20-30 where the learning processes split into two distinct groups. In this figure, a more uniform behaviour of QRNG methods are noted, unlike the previous experiment. The behaviours of PRNG distributed models are extremely erratic and in some cases, very slow in terms of improvements made. Fig. \ref{endgraphmental} show a higher resolution view of the data in terms of the end of the learning process when terminated at epoch 100, a clear distinction of results can be seen and a concrete separation can be drawn between the two groups of models except for two intersecting processes. It should be noted that by this point, the learning process has not settled towards a true best fitness, but a vast and clear separation has occurred.

For PRNG, the standard deviation between all 25 results was 0.98. The standard deviation for QRNG was somewhat smaller at 0.74. The mean of all results was 63.84\% for PRNG distributions and 66.45\% for QRNG distribution, a slightly superior result. The maximum classification accuracy achieved by the PRNG initial distribution was 65.35\% whereas QRNG achieved a somewhat higher best result of 68.17\%. The worst-best result for PRNG distribution networks was 62.28\%, and was 65.31\% for QRNG distribution networks. For this problem, the differences between the initial distribution of PRNG and QRNG weights are noticeable, QRNG distribution results are consistently better than PRNG approaches to initial weight distribution.

\subsection{Random Tree and Quantum Random Tree Classifiers}
Experiments 2a and 2b make use of the same datasets as in 1a and 1b respectively. In this experiment, 200 Random Tree classifiers are trained for each dataset. These are, again, comprised of two sets; firstly 100 Random Tree (RT) classifiers which use Pseudorandom numbers, and secondly, 100 \textit{Quantum Random Tree} (QRT) classifiers, which source their random numbers from the QRNG. Random Numbers are used to select the n-random attribute subsets at each split. 

\subsubsection{Accent Classification}
\begin{figure}
    \centering
    \includegraphics[scale=0.5]{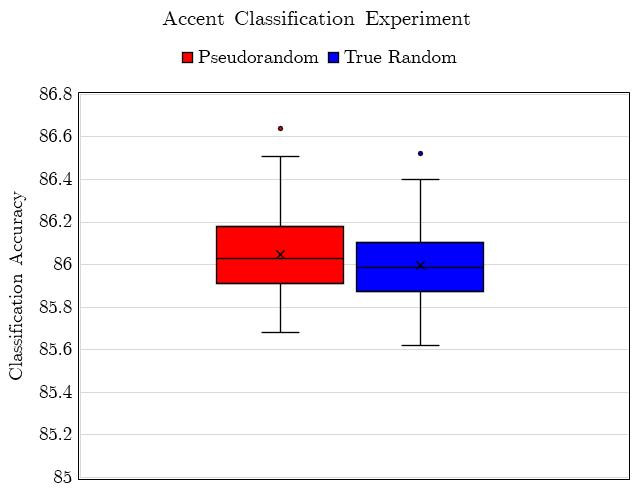}
    \caption{A Comparison of results from 200 Random Tree Classifiers, 100 using PRNG and 100 using QRNG on the Accent Classification Dataset}
    \label{randomtreeaccent}
\end{figure}

200 Experiments are graphically represented as a box-and-whisker in Fig. \ref{randomtreeaccent}. The most superior classifier was the RT with a best result of 86.64\% and worst of 85.68\%, on the other hand, the QRT achieved a best accuracy of 86.52\% and worst of 85.62\%. Best and worst results of the two models are extremely similar. The standard deviation of results of the RT was 0.19 and the QRT similarly had a standard deviation of 0.17. The range of the RT results was 0.96 and QRT results had a similar range of 0.9. Interestingly, a similar pattern is not only found in results, but also with the high outlier too when considered relative to the model's median point. Though an overall slight superiority is seen in pseudorandom number generation, the two models are considerably similar in their abilities. 

\subsubsection{Mental State Classification}
\begin{figure}
    \centering
    \includegraphics[scale=0.5]{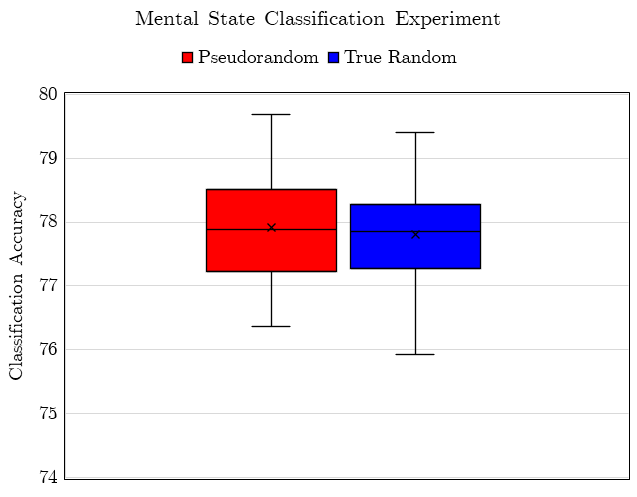}
    \caption{A Comparison of results from 200 Random Tree Classifiers, 100 using PRNG and 100 using QRNG on the Mental State EEG Dataset}
    \label{randomtreemental}
\end{figure}

Fig. \ref{randomtreemental} shows the distribution for the 200 Random Tree classifiers trained on the Mental State dataset. The standard deviation of results from the RT was 0.81 whereas it was slightly lower for QRT at 0.73. The best result achieved by the RT was 79.68\% classification accuracy whereas the best result from the QRT was 79.4\%. The range of results for RT and QRT were a similar 3.31 and 3.47 respectively. Overall, very little difference between the two models occurs. The distribution of results can be seen to be extremely similar to the first RT/QRT experiment when compared to Fig. \ref{randomtreeaccent}.

\subsection{Random Forest and Quantum Random Forest Classifiers}
In this third experiment, the datasets are classified using two models. Random Forests (RF) which use a committee of Random Trees to vote on a Class, and Quantum Random Forests (QRF) which use a committee of Quantum Trees to vote on a class. For each dataset, 10 of these models are trained, with a committee of 10 to 100 Trees respectively. 

\subsubsection{Accent Classification}
\begin{figure}
    \centering
    \includegraphics[scale=0.955]{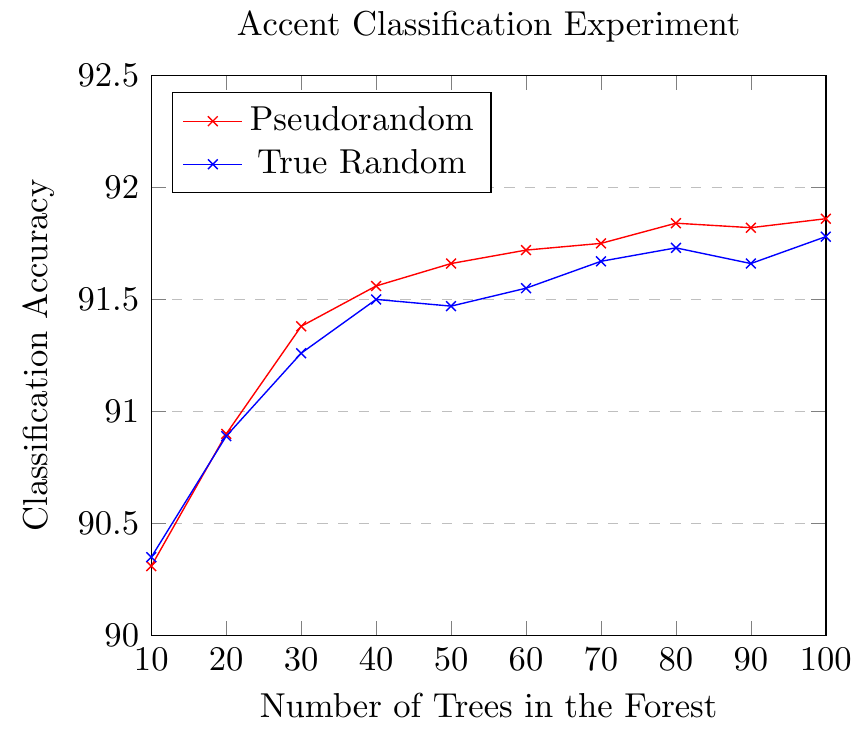}
\caption{Classification Accuracies of 10 Random Forest and 10 Quantum Forest Models on the Accent Classification Dataset}
\label{accentforest}
\end{figure}
%
%
%
%

The results from the Accent Classification dataset for the RF and QRF methods can be observed in Fig. \ref{accentforest}. The most superior models both used a committee of 100 of their respective trees, scoring two similar results of 91.86\% with Pseudo-randomness and 91.78\% for Quantum randomness. Standard deviation of RF results are 0.5\% whereas QRF has a slightly lower deviation of 0.43. The worst result by RF was 90.31\% classification accuracy at 10 Random Trees, the worst result by the QRF was similarly 10 Quantum Trees at 90.36\% classification accuracy (+0.05). The range of RF results was 1.55, compared to the QRF results with a range of 1.43.

\subsubsection{Mental State Classification}
\begin{figure}
\centering
\includegraphics[scale=0.955]{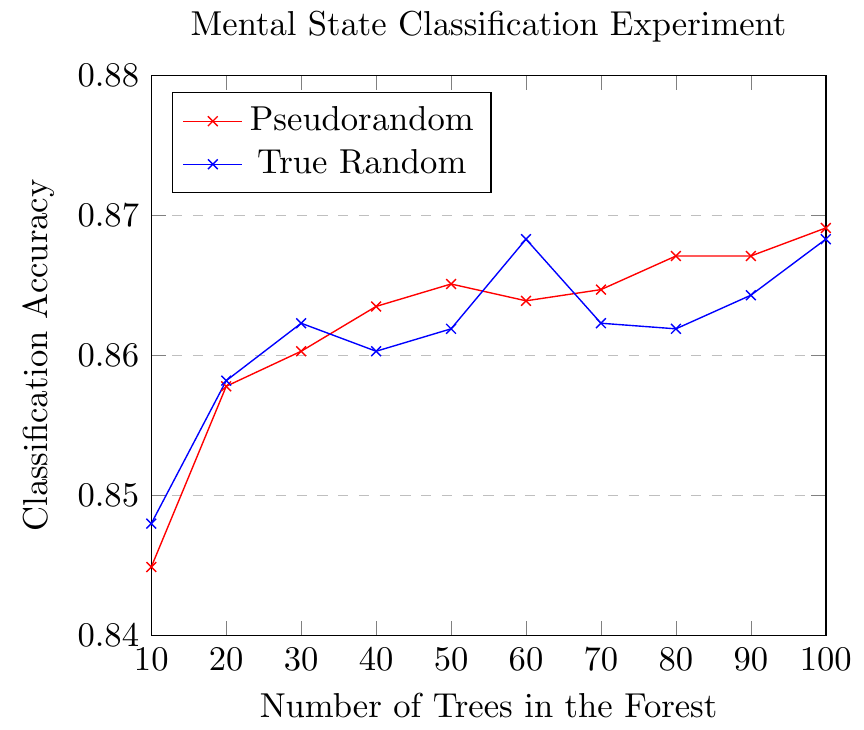}
%
%
%
%
%
%
%
\caption{Classification Accuracies of 10 Random Forest and 10 Quantum Forest Models on the EEG Mental State Classification Dataset}
\label{mentalforest}
\end{figure}

The results from the Mental State EEG Classification dataset for the RF and QRF methods can be observed in Fig. \ref{mentalforest}. The most superior model for the RF was 86.91\% with a committee of 100 trees whereas the best result for QRF was 86.83\% achieved by committees of both 100 and 60 trees. The range of QRF results were slightly lower than that of the RF, measured at 2.34 and 2.42 respectively. Although initially considered negligible, this same pattern was observed in the previous experiment in Fig. \ref{accentforest}. Additionally, the standard deviation of RF was higher at 0.69 compared to 0.65 in QRF. \\

Though very similar results were produced, the first QRF best result required approximately 60\% of the computational resources to achieve compared to the best RF result. Unlike the first Forest experiment, the patterns of the two different models are vastly different and often alternate erratically. This suggests somewhat that the two models should both be benchmarked in order to increase the chances of discovering a more superior model, considering the level of data dependency on the classification accuracies of the models.

\subsection{CNN: Initial Random Weight Initialisation for Computer Vision}
Experiment 4a and 4b make use of the MNIST and CIFAR-10 image datasets respectively. In 4a, an ANN is initialised following the same PRNG and QRNG methods utilised in Experiment 1 and trained to classify the MNIST handwritten digits dataset. In 4b, the final dense layer of the CNN are initiliased through the same methods.  

\subsubsection{MNIST Image Classification}
For the purpose of scientific recreation, the architecture for MNIST classification is derived from the official Keras example\footnote{https://github.com/keras-team/keras/tree/master/examples}. This is given as two sets of two identical layers, a hidden layer of 512 densely connected neurons followed by a dropout layer of 0.2 to prevent over-fitting. All hidden neurons, as with other experiments in this study, are initialised randomly within the standard -0.5 to 0.5 range. 25 of these are generated by a PRNG and the other 25 by a QRNG, producing observable results of 50 models in total. 

\begin{figure}
    \centering
    \includegraphics[scale=0.55]{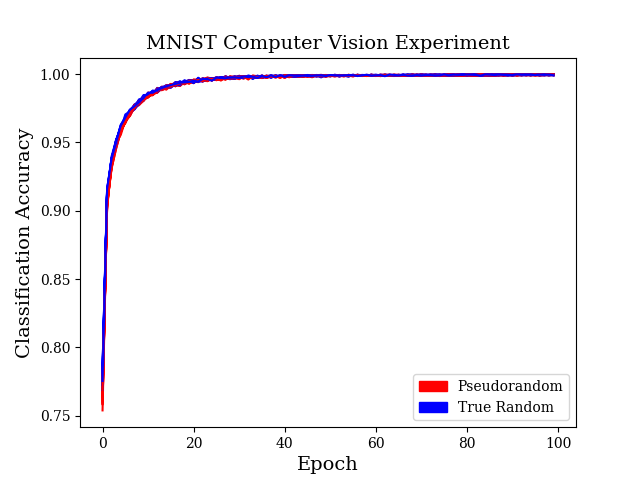}
    \caption{The Full Learning Process of 50 Deep Neural Networks, 25 with PRNG and 25 with QRNG Initially Distributed Weights in MNIST Image Dataset Classification}
    \label{mnistgraph}
\end{figure}

\begin{figure}
    \centering
    \includegraphics[scale=0.55]{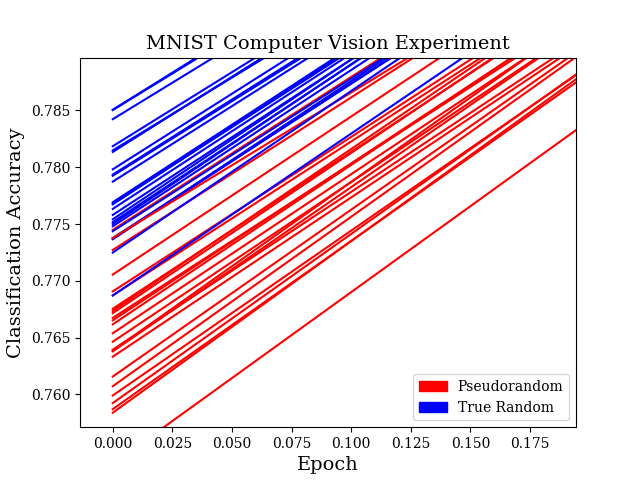}
    \caption{Initial (pre-training) Classification Abilities of 50 Deep Neural Networks, 25 with PRNG and 25 with QRNG Initially Distributed Weights in MNIST Image Dataset Classification}
    \label{mnistinitial}
\end{figure}

\begin{figure}
    \centering
    \includegraphics[scale=0.55]{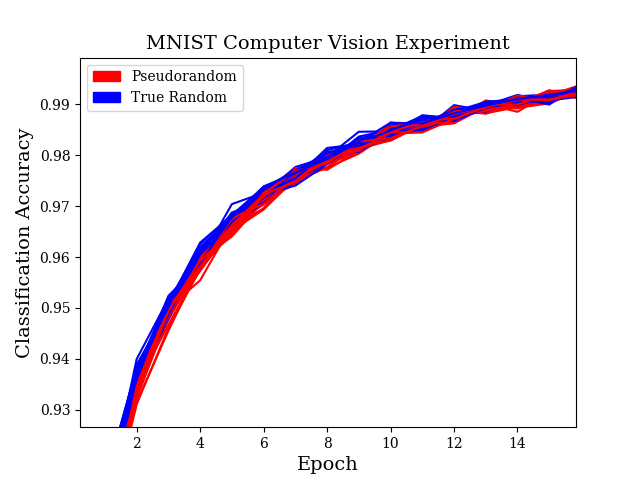}
    \caption{The Initial Learning Curve Experienced for 50 Deep Neural Networks, 25 with PRNG and 25 with QRNG Initially Distributed Weights in MNIST Image Dataset Classification}
    \label{mnistcurve}
\end{figure}

Due to the concise nature and close results observed in the full process showed in Fig. \ref{mnistgraph}, two additional graphs are presented; firstly, the graph in Fig. \ref{mnistinitial} shows the classification abilities of the models before any training occurs. Within this, a clear distinction can be made, the starting weights generated by QRNG are almost exclusively superior to those generated by PRNG, providing the QRNG models with a superior starting point for learning. The distinction continues to occur throughout the initial learning curve, observed in Fig. \ref{mnistcurve}, not too dissimilar to the results in the previous experiment. At the pre-training abilities of the two methods of weight initialisation, dense areas can be observed at approx 77.5\%  Finally, at around epochs 10-14, the resultant models begin to converge and the separation becomes less prominent. This is shown through both sets of models having identical best classification accuracies of 98.64\%m suggesting a true best fitness may possibly have been achieved. Worst-best accuracies are also indistinguishably close, 98.27\% for QRNG models and 98.25\% for PRNG models, population fitnesses are extremely dense and little entropy exists throughout the whole set of final results.

\subsubsection{CIFAR-10 Image Classification}
In the CNN experiment, the CIFAR-10 image dataset is used to train a Convolutional Neural Network. The two number generators are applied for the initial random weight distribution of the final hidden dense layer, after feature extraction has been performed by the CNN operations. The network architecture is constructed as is the official Keras Development Team example for Scientific purposes in ease of recreation of the experiment. In this architecture, one hidden dense layer of 512 units precedes the final classification output, and weights are generated within the bounds of -0.5 and 0.5 as is a standard in neural network generation. 50 CNNS are trained, all of which are structurally identical except for that 25 have their dense layer weights initialised by PRNG and the other 25 have their dense layer weights initialised by QRNG. 

\begin{figure}
    \centering
    \includegraphics[scale=0.55]{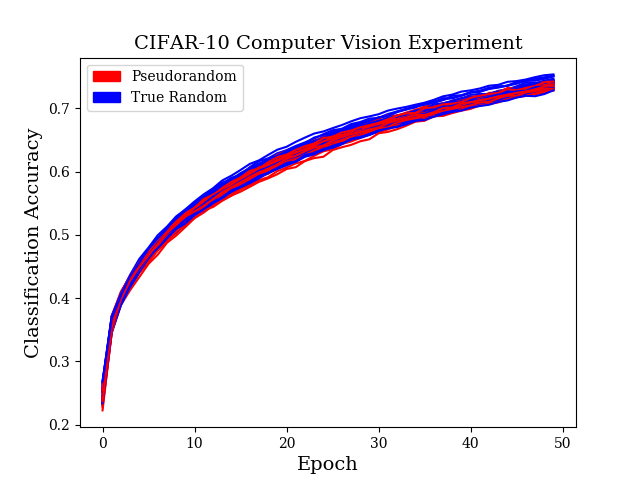}
    \caption{The Full Learning Process of 50 Convolutional Neural Networks, 25 with PRNG and 25 with QRNG Initially Distributed Weights for the Final Hidden Dense Layer in CIFAR-10 Image Dataset Classification}
    \label{cifargraph}
\end{figure}

\begin{figure}
    \centering
    \includegraphics[scale=0.55]{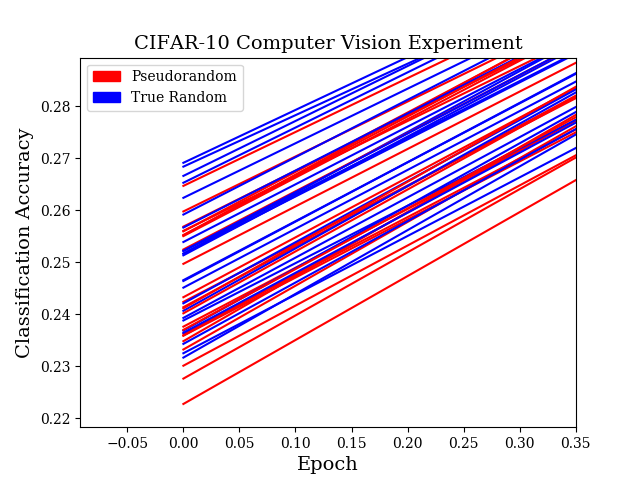}
    \caption{Initial (pre-training) Classification Abilities of 50 Convolutional Neural Networks, 25 with PRNG and 25 with QRNG Initially Distributed Weights for the Final Hidden Dense Layer in CIFAR-10 Image Dataset Classification}
    \label{cifarinitial}
\end{figure}

\begin{figure}
    \centering
    \includegraphics[scale=0.55]{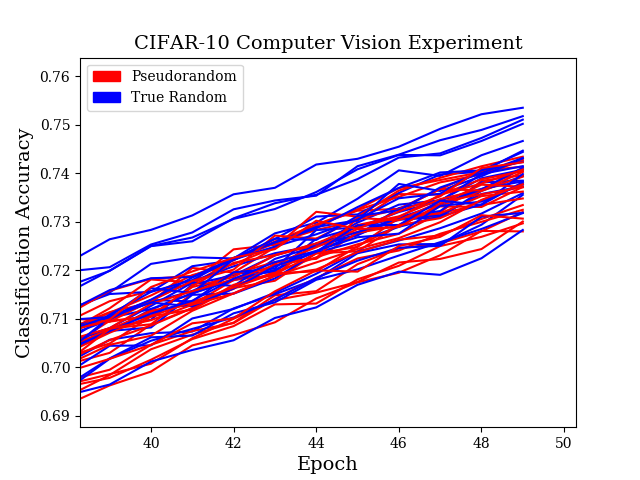}
    \caption{The Learning within the Final Epochs for 50 Convolutional Neural Networks, 25 with PRNG and 25 with QRNG Initially Distributed Weights for the Final Hidden Layer in CIFAR-10 Image Dataset Classification}
    \label{cirfarend}
\end{figure}

Fig. \ref{cifargraph} shows the full learning process of the two different methods of initial weight distribution. It can be observed that there are roughly three partitions of results between the two methods, the pattern is visually similar to the ANN learning curve in the MNIST Computer Vision experiment. Fig \ref{cifarinitial} shows the pre-training classification abilities of the initial weights, distribution is relatively equal and unremarkable unless compared to the final results of the training process in Fig. \ref{cirfarend}; the four best initial distributions of network weights, all are of that which have been generated by the QRNG, continue to be the four superior overall models. It must be noted although, that the rest of the models regardless of RNG method, are extremely similar and no other divide is seen by the end of the process. \\

The six overall most superior models were all initialised by QRNG, the best result being a classification accuracy of 75.35\% at epoch 50. The seventh best model was the highest scoring model that had dense layer weights initialised by PRNG, scoring a classification accuracy of 74.43\%. The worst model produced by the QRNG was that which had a classification accuracy of 71.91\%, slightly behind this was the overall worst model from all experiments, a model initialised by the PRNG with an overall classification ability of 71.82\%. The QRNG initialisation therefore outperformed PRNG by 0.92 in the best case, and outperformed PRNG by 0.09 in the worst case. The average result from both methods of distribution. The average result between the two models was equal, at 73.3\% accuracy. \\

It must be noted that by epoch 50 the training process was still producing increasingly better results, but computational resources available limited the 50 networks to be trained for this amount of time.

\section{Future Work} \label{futurework}
It was observed in those experiments that did stabilise, results as expected reached closer similarities. With resources, future work should concern the further training of models to observe this pattern with a greater reach of examples. Extensive computational resources would be required to train such an extensive amount of networks. \\

Furthermore, the patterns in Fig. \ref{mentalforest}, Quantum vs Random Forest for Mental State Classification, suggest that the two forests have greatly different situational classification abilities and may produce a stronger overall model if both used in an ensemble. This conjecture is strengthened through a preliminary experiment; a vote of maximum probability between the two best models in this experiment (QF(60) and RF(100)) produces a result of 86.96\% which is a slight, and yet superior classification ability. The forests ensembled with other forests of their on type on the other hand do not improve. With this discovery, a future study should consider ensemble methods between the two for both deriving a stronger overall classification process, as well as to explore the patterns in the ensemble of QRNG and PRNG based learning techniques. This, at the very minimum, would require the time and computational resources to train 100 models to explore the two sets of ten models produced in the related experiment, though exploring beyond this, or even a full bruteforce of each model increasing their population of forests by 1 rather than 10 would produce a clearer view of the patterns within.\\

Of the most noticeable effects of QRNG and PRNG in machine learning, many of the neural network experiments show greatly differing patterns in learning patterns and their overall results when using PRNG and QRNG methods to generate the initial weights for each neuron within hidden layers. Following this, further types of neural network approaches should be explored to observe the similarities and differences that occur. In addition to this, the architectures of networks are by no means at an optimum, the heuristic nature of the network should also be explored, by techniques such as a genetic search, for it too requires the idea of random influence ~\cite{jordanevoalgmlp, bird2019devo}.

\section{Conclusion} \label{conclusion}
To conclude, this study performed 8 individual experiments to observe the effects of Quantum and Pseudorandom Number Generators when applied to multiple machine learning techniques. Some of the results were somewhat unremarkable as expected, but some effects presented profound differences between the two, many of which are as of yet greatly unexplored. Based on these effects, possibilities of future work has been laid out in order to properly explore them. \\

Though observing superposition provides perfectly true randomness, this also provides a scientific issue in the replication of experiments since results cannot be coerced in the same nature a PRNG can through a seed. In terms of cybersecurity, this nature is ideal ~\cite{yang2014quantum, stipcevic2012quantum}, but provides frustration in a research environment since only generalised patterns at time \textit{t} can be analysed ~\cite{svore2016quantum}. This is overcome to an extent by the nature of repetition in the given experiments, many countless classifiers are trained to provide a more average overview of the systems.\\

The results for all of these experiments suggest that data dependency leads to no concrete positive or negative effect conclusion for the use of QRNG and PRNG since there is no clear superior method. Although this is true, pseudo-randomness on modern processors are argued to be indistinguishable from true randomness, but clear patterns have emerged between the two. The two methods do inexplicably produce different results to one another when employed in machine learning, an unprecedented, and as of yet, relatively unexplored line of scientific research. In some cases, this was observed to be a relatively unremarkable, small, and possibly coincidental difference; but in others, a clear division separated the two. \\

The results in this study are indicative of a profound effect on patterns observed in machine learning techniques when random numbers are generated either by the rules of classical or quantum physics. Their effects being positive or negative are seemingly dependent on the data at hand, but regardless, the fact that two methods of randomness ostensibly cause such disparate effects juxtapose to the current scientific processes of their usage should not be underestimated. Rather, it should be explored.

\section*{Appendices}
\section*{1 Quantum Assembly Language for Random Number Generation}
Note: Code comments (\#) are \textbf{not} Quantum Assembly Language and are simply for explanatory purposes. The following code will place a quanta into superposition via the Hadamard gate and then subsequently measure the state and store the observed value. The state is equally likely to be observed at either 1 or 0.

\begin{lstlisting}
#Electron zero to Hadamard Gate
H 0 
#Declare memory space 'ro' of one bit
DECLARE ro BIT[1] 
#Measure the qubit at 0th index of 'ro'
MEASURE 0 ro[0] 
\end{lstlisting}

\section*{2 Python Code for Generating a String of Random Bits}
The following code generates a random 32-bit integer by observing an electron in superposition which produces a true random result of either 1 or 0. The result is amended at each individual observation until 32 bits have been generated. Decimal conversion takes place and two files are generated; a raw text file containing the decimal results and a CSV containing a column of binary integers and their decimal equivalents. 

\begin{lstlisting}
from pyquil.quil import Program
from pyquil.gates import H

# Select the lattice of Qubits
lattice = "Aspen-1-5Q-B"  
# Initialise QPU
qpu = get_qc(lattice) 

#Place electron 0 into superposition
numbers = Program(H(0)) 
#Observe the superposition
getNum = numbers.measure_all() 
#Print the Quantum Assembly Language
print(getNum)

compiled_program = qpu.compile(numbers)

#Length of integer to generate
numbers = 32 
#How many integers to generate
toGenerate = 1 

print("\n Random number of " + str(numbers) + " bits:")


for y in range(0, 10000): 
	output = ""
	for x in range(0, toGenerate):
		
		#Run the code on a Quantum Processing Unit
		result = qpu.run(compiled_program)
		#Observe the superposition
		result = result[0][0] 
		
		output += str(result)

	print("\n\n Random no." +  str(y) + " is: " + output)
	decimal = int(output, 2)

	with open("numbers.txt", "a") as myfile:
		myfile.write("\n" + str(decimal))
		
	with open("random.csv", "a") as myfile:
		myfile.write("\n" + str(output) + "," + str(decimal))
\end{lstlisting}

\section*{Acknowledgments}
The authors would like to thank \textit{Rigetti Computing} for granting access to their Quantum Computing Platform.


\bibliographystyle{spbasic}      

\end{document}